\documentclass[10pt,twocolumn,letterpaper]{article}

\usepackage[pagenumbers]{iccv} 
\usepackage{marvosym}
\definecolor{iccvblue}{rgb}{0.21,0.49,0.74}
\usepackage[pagebackref,breaklinks,colorlinks,allcolors=iccvblue]{hyperref}

\newcommand{\ModelName}{\emph{FullDiT}}
\newcommand{\BenchmarkName}{\emph{FullBench}}
\usepackage{threeparttable}
\usepackage{multirow} 
\usepackage{float}
\definecolor{brightmaroon}{rgb}{0.76, 0.13, 0.28}

\def\paperID{1415} 
\def\confName{ICCV}
\def\confYear{2025}

\title{
FullDiT: Multi-Task Video Generative Foundation Model with Full Attention
}

\author{
Xuan Ju$^{12}$\thanks{Work done during internship at Kuaishou Technology} \quad Weicai Ye$^{1\textrm{\Letter}}$ \quad
  Quande Liu$^{1}$ \quad
  Qiulin Wang$^1$ \quad
  Xintao Wang$^1$ \quad \\
  Pengfei Wan$^{1}$ \quad
  Di Zhang$^1$ \quad
  Kun Gai$^1$ \quad Qiang Xu$^{2\textrm{\Letter}}$\\
  \textnormal{$^1$Kuaishou Technology} \qquad \textnormal{$^2$The Chinese University of Hong Kong} 
  \\
  \textbf{\small \urlstyle{tt}{\url{https://fulldit.github.io}}}
}

\begin{document}

\twocolumn[{
\maketitle
\vspace{-5mm}
\centerline{
\includegraphics[width=0.99\linewidth]{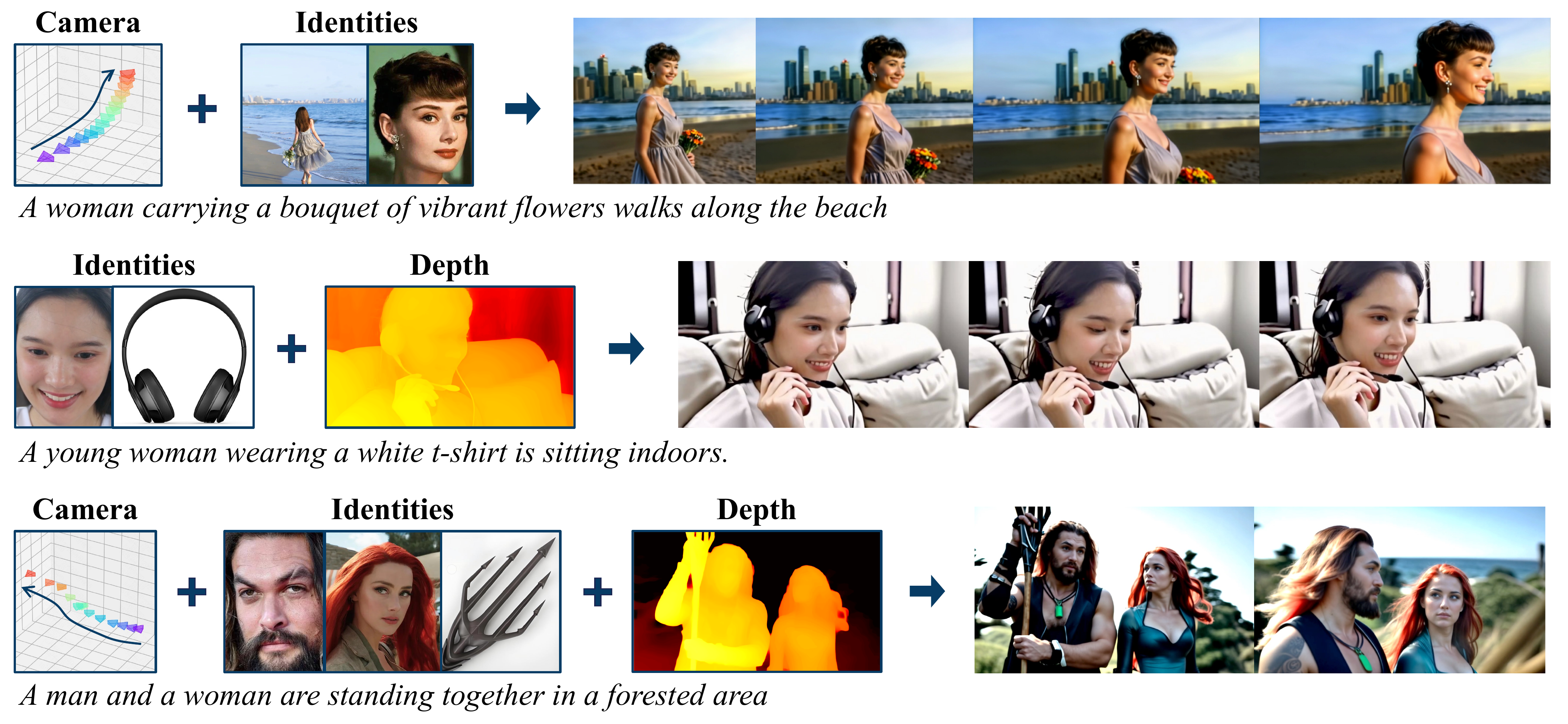}}
\vspace{-2mm}
\captionof{figure}
{
\textbf{\ModelName~is a multi-task video generative foundation model that unifies conditional learning with full self-attention}. With self-attention’s long-context learning ability, \ModelName~can flexibly take different combinations of input to generate high-quality videos.
}
\vspace{5mm}
\label{fig:teaser}
}]
\def\thefootnote{$\ast$}\footnotetext{Work done during internship at Kuaishou Technology.}\def\thefootnote{\textrm{\Letter}}\footnotetext{Corresponding Author.}

\begin{abstract}

\vspace{-1mm}
Current video generative foundation models primarily focus on text-to-video tasks, providing limited control for fine-grained video content creation. Although adapter-based approaches (e.g., ControlNet) enable additional controls with minimal fine-tuning, they encounter challenges when integrating multiple conditions, including: \emph{branch conflicts} between independently trained adapters, \emph{parameter redundancy} leading to increased computational cost, and \emph{suboptimal performance} compared to full fine-tuning. To address these challenges, we introduce \ModelName, a unified foundation model for video generation that seamlessly integrates multiple conditions via unified full-attention mechanisms. By fusing multi-task conditions into a unified sequence representation and leveraging the long-context learning ability of full self-attention to capture condition dynamics, FullDiT reduces parameter overhead, avoids conditions conflict, and shows scalability and emergent ability. We further introduce \BenchmarkName~for multi-task video generation evaluation. Experiments demonstrate that FullDiT achieves state-of-the-art results, highlighting the efficacy of full-attention in complex multi-task video generation.

\end{abstract}

\vspace{-4mm}
    
\section{Introduction}
\label{sec:introduction}

\vspace{-1mm}

The pre-training of video generative foundation models has predominantly adhered to a paradigm focused solely on text-to-video generation, benefiting from its simplicity and broad applicability. However, relying solely on textual prompts offers insufficient granularity, failing to provide precise and direct manipulation over critical video attributes. Real-world creative industries—such as filmmaking, animation, and digital content creation—frequently require fine-grained control across multiple aspects of generated videos, such as camera movements, character identities, and scene layout. To meet these multifaceted demands, recent works (e.g., ControlNet~\cite{controlnet} and T2I-Adapter~\cite{t2i_adapter}) typically incorporate additional control signals via adapter-based methods, where adapter networks process supplementary conditions separately and integrate them through mechanisms like cross-attention~\cite{animateanyone} or addition~\cite{cameractrl} operations. These adapter-based methods have gained popularity primarily due to their minimal parameter tuning, enabling rapid deployment and flexibility in single-task scenarios.

\begin{figure*}[htbp]
    \centering
    \includegraphics[width=0.99\linewidth]{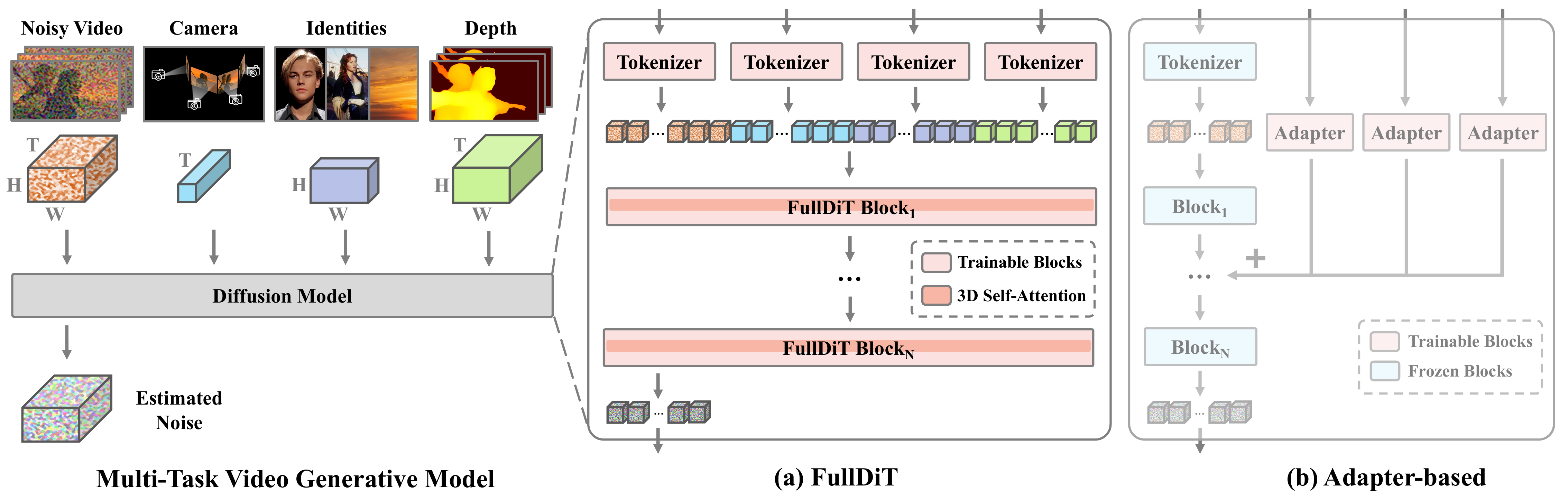}
    \vspace{-2mm}
    \caption{\textbf{Overview of \ModelName~architecture and comparison with adapter-based models.} We present the diffusion process of the multi-task video generative model on the left. For research purposes, this paper shows input conditions consisting of \textbf{temporal-only} cameras, \textbf{spatial-only} identities, and \textbf{temporal-spatial} depth video. Additional conditions can be incorporated into this model architecture for broader applications. Shown in (a),  \ModelName~unifies various inputs with procedures: (1) patchify and tokenize the input condition to a unified sequence representation, (2) concat all sequences together to a longer one, and (3) learn condition dynamics with full self-attention.
    By comparison, earlier adapter-based approaches (shown in (b)) use distinct adapter designs that operate independently to process various inputs, leading to branch conflicts, parameter redundancy, and suboptimal performance.
    Each block's subscript indicates its layer index.
    } 
    \vspace{-5mm}
\label{fig:model}
\end{figure*}

Although these adapter-based techniques have shown promise in single-task scenarios, extending them to multimodal and multi-condition video generation scenarios exposes significant drawbacks. Firstly, adapters trained independently can clash when combined (termed ``\emph{branch conflicts}"), resulting in compromised overall generation performance~\cite{humansd}. Secondly, these condition-specific adapters often introduce parameter redundancy~\cite{ctrlora}. Finally, adapters usually achieve suboptimal generation quality compared to methods that fine-tune the entire model~\cite{deltatuning,humansd}. These limitations indicate a clear gap and an urgent need for a more robust and integrated solution capable of effectively addressing multiple conditions simultaneously.

In response to these challenges, this paper presents \ModelName, a novel video generative foundation model that harnesses a unified full-attention framework to integrate diverse condition signals. 
As shown in Figure~\ref{fig:model}, unlike adapter-based approaches that introduce separate processing branches, 
\ModelName~integrates multiple conditions into a single coherent sequential representation and learns the mapping from conditions to video with full self-attention. 
Our key insight is that unified attention facilitates powerful cross-modal representation learning, effectively capturing complex temporal and spatial correlations. By jointly processing conditions within a shared attention module, \ModelName~inherently resolves branch conflicts common in adapter-based methods, reduces parameter redundancy by avoiding separate adapters, and achieves superior multi-task integration through effective end-to-end training.

Furthermore, our solution enables scalable extension to additional modalities or conditions without major architectural modifications. As training data volume increases, \ModelName~exhibits strong scalability and reveals emergent capabilities, successfully generalizing to previously unseen combinations of conditions, \eg, simultaneously controlling camera trajectories and character identities.

During pre-training, we observed that more challenging tasks require additional training and should be introduced earlier. 
Conversely, introducing easier tasks too early may cause the model to prioritize learning them, hindering the convergence of challenging tasks. 
Based on this observation, we follow the training order of text, camera, identity, and depth.
Afterward, we conduct quality fine-tuning~\cite{emu} with manually selected high-quality data to enhance the video's dynamics, controllability, and visual quality.

Existing video generation benchmarks focus primarily on single-task evaluations, which makes them inadequate for evaluating models that integrate multiple different conditions. To address this critical gap, we introduce \BenchmarkName, the first benchmark specifically designed to evaluate multi-condition video generation tasks comprehensively, consisting of 1,400 carefully curated cases covering various condition combinations
, enabling a robust assessment of multi-task video generative capabilities.

Our contributions are highlighted as follows:
\begin{itemize}
    \item We introduce \ModelName, the first video generative foundation model that leverages unified full attention mechanisms for integrating multiple control signals. 
    \item We propose a progressive training strategy for multi-task generation, demonstrating that a tailored condition training order leads to better convergence and performance.
    \item We construct and release \BenchmarkName, the first benchmark designed explicitly for evaluating multi-condition video generation, filling an evaluation gap.
\end{itemize}

Through extensive experimentation, we demonstrate that \ModelName~achieves state-of-the-art performance across multiple video generation tasks, showcasing emergent capabilities in combining diverse, previously unseen tasks.

\section{Related Work}\label{sec:related_work}

\subsection{Video Generation}

Video generation has progressed significantly in recent years, transitioning from early GAN-based models~\cite{vgan,mcvd} to the latest diffusion models~\cite{sora,moviegen}. 
Due to the scalability and effectiveness of the transformer architecture~\cite{dit}, the exploration of video diffusion models has gradually shifted from convolution-based architectures~\cite{makeavideo,alignyourlatents,emuvideo,videocrafter2,moonshot} to transformer-based architecture~\cite{walt,miradata,snapvideo,sora,moviegen}.
The most common practice is to divide the video into small patches and then feed a sequence of patches into a full-attention transformer architecture. 
Control information, such as text, is injected into the model via cross-attention or adapters.

The majority of the previous video generative foundation models are pre-trained with pure text condition and use cross-attention for its injection. In pursuit of more fine-grained and user-customized video control, numerous controllable video generation methods focus on the control of motion~\cite{vd3d,cavia,cameractrl,motionctrl,motionclone,cami2v,motionprompting}, identity~\cite{animateanything,animateanyone,consisid,dreamvideo,idanimator,magicme,motionbooth,conceptmaster,VideoAlchemist,MagicMirror}, and structure~\cite{trackgo,sparsectrl,ctrladapter,videocontrolnet,controlvideo,videocomposer} have been introduced, most of which are based on adapter-based frameworks. However, previous studies~\cite{deltatuning,humansd} have shown that adapter-based methods tend to produce suboptimal results with larger model parameter sizes. Furthermore, they are not well-adapted for multi-task generation because of mutual conflicts and the problem of parameter wastage.

Following the proposal of MMDiT in Stable Diffusion 3~\cite{sd3}, it was recognized that conditions do not have to be input with cross-attention, but can also be fused using full self-attention~\cite{hunyuanvideo}. 
However, no further exploration has been made into applying the same unified full-attention framework to introduce multiple controls in video generation, which is precisely the focus of this paper.

\subsection{Multi-Task Image and Video Generation}

Previous visual generation models are mainly diffusion-based, while language generation models primarily rely on autoregressive next-token prediction. Following these two strategies, unified multi-task image and video generation can be categorized into two main streams as follows:

\noindent \textbf{Autoregressive Models.} With the development of multimodal language models with a paradigm of autoregressive next-token prediction, there is a growing interest in exploring ways to incorporate visual generation into these models. 
Recent efforts can be categorized into two main paradigms: one leverages discrete tokens for image representation, treating image tokens as part of the language codebook~\cite{unifiedio,chameleon,luminamgpt,janus}, while the other uses continuous tokens, combining diffusion models with language models to generate images~\cite{transfusion,showo,omnigen,seedx}. There have also been efforts to explore video generation using discrete tokens~\cite{videopoet,emu3}. 
While these efforts to incorporate generation into multimodal language models are highly valuable, the prevailing consensus remains that diffusion is currently the most effective approach for video generation.

\noindent \textbf{Diffusion Models.} Although diffusion models are very effective in generating images and videos, the exploration of multi-task image and video generation is still in the early stages.
One Diffusion~\cite{onediffusion} and UniReal~\cite{unireal} employ 3D representations to combine visual conditions, enabling unified image generation and editing tasks.
OmniControl~\cite{omnicontrol,groupdiffusion} uses the MMDiT~\cite{sd3} design and adds additional branches along with the noise branch and the text branch to incorporate other conditions. 
OmniFlow~\cite{omniflow} further introduces the understanding and generation of audio and text using a similar MMDiT design.
For video generation, the incorporation of temporal information introduces additional challenges in multitask video generation.
While OmniHuman~\cite{omnihuman} has explored human video generation with multiple controls, it incorporates various types of condition addition methods (\textit{i.e.}, cross-attention, and channel concatenation). Since videos are typically treated as sequential structures in contemporary generation backbones, we aim to explore whether a unified sequential input form could effectively integrate multimodal video condition control. Although a few of previous works~\cite{unireal} have explored using unified sequential input for image generation, we highlight three key differences compared to video generation: 
(1) All the conditions they incorporated are confined within the image modality. Thus, they do not offer guidance on how to combine conditions from different distributions or validate the generalization ability. (2) Their focus is on image generation, without considering temporal information. (3) They did not investigate the emergent abilities arising from the interaction among conditions and the scalability of training data. \ModelName~represents the first step in exploring them.

\vspace{-2mm}
\section{Method}

In this section, we present the details of our proposed framework, \ModelName. The goal of \ModelName~is to utilize multiple conditions (\textit{e.g.}, text, camera, identities, and depth) to generate high-quality videos with fine-grained control. While this study focuses on a limited set of conditions, the method can be adapted and expanded to various conditions.

\subsection{Preliminary}

Video diffusion models~\cite{sora,moviegen} learn the conditional distribution $p(\mathbf{x}|C)$ of video data $\mathbf{x}$ given conditions $C$. In the diffusion process with the formulation of Flow Matching~\cite{flowmatching}, noise sample $\mathbf{x_0} \sim \mathcal{N}(0,1)$ is progressively transited into clean data $\mathbf{x_1}$ with $\mathbf{x_t} = t\mathbf{x_1} + (1-(1-\sigma _{min})t)\mathbf{x_0}$, where $\sigma _{min} = 10^{-5}$ and timestep $t \in [0, 1]$. The learnable model $u$ is trained to predict the velocity $V_t=\frac{d\mathbf{x_t}}{dt}$, which can be further derived as:

\vspace{-2mm}
$$
V_t=\frac{d\mathbf{x_t}}{dt} = \mathbf{x_1} - (1-\sigma _{min}) \mathbf{x_0}
$$

Thus, the model $u$ with the parameter $\Theta$ is optimized by minimizing the mean squared error loss $\mathcal{L}$ between the ground truth velocity and the  model prediction:

\vspace{-6mm}
$$
\mathcal{L}=\mathbb{E}_{t,\mathbf{x_0},\mathbf{x_1},C}\left \| u_{\Theta}(x_t,t,C) - (\mathbf{x_1} - (1-\sigma _{min}) \mathbf{x_0})  \right \| 
$$

During the inference process, the diffusion model first samples $\mathbf{x_0} \sim \mathcal{N}(0,1)$, then uses an ODE solver with a discrete set of $N$ timesteps to generate $\mathbf{x_1}$.

\subsection{Overview}
\label{sec:overview}

We illustrate the comparison of \ModelName~with previous adapter-based frameworks in Figure~\ref{fig:model}. For adapter-based condition insertion methods (as shown in Figure~\ref{fig:model} (b)), an additional adapter is required for each condition. 
This leads to design complexity and increased parameter overhead, as each condition requires a specialized design module for feature processing, resulting in limited scalability for introducing new conditions.
Moreover, since adapters are trained independently for each task without sharing information, arbitrary integration may cause conflicts and degrade overall performance. 
Compared to adapter-based methods, \ModelName~directly merges all conditions at an early stage (as shown in Figure~\ref{fig:model} (a)), \textit{i.e.}, each condition is tokenized into a sequence representation and is subsequently concatenated and fed into the transformer blocks. 
This facilitates a more thorough fusion among conditions without the need of additional parameters for each condition.

Following previous works~\cite{opensoraplan,moviegen}, \ModelName~adopts a transformer architecture comprising 2D self-attention, 3D self-attention, cross-attention, and feedforward networks\def\thefootnote{1}\footnotetext{Due to the constrain of pre-trained text-to-video model, cross-attention is employed to incorporate text information, while 2D self-attention is utilized to reduce computational overhead. But ideally, the model should depend solely on 3D full attention and feedforward layers.}\footnotemark[1]. More specifically, \ModelName~first tokenizes video, camera, identities, and depth conditions into sequence representations by patchifying them and then mapping them to the hidden dimension using one layer of convolution. Afterward, in each \ModelName~block, the sequence latent first passes 2D self-attention with 2D rotational position encoding (RoPE) to enhance spatial information learning. Then, the latent passes 3D self-attention with 3D RoPE that enables joint modeling of spatial and temporal information among multiple conditions. This allows for a natural interaction among different input signals in both spatial and temporal, thus ensuring optimal performance. 
Meanwhile, diffusion timesteps are mapped via AdaLN-Zero to four sets of scale, shift, and gate parameters, which are subsequently injected into the 2D self-attention, 3D self-attention, cross-attention, and feedforward layers.

Given a set of conditions $\{C_i | i=1\cdots n\}$, our goal is to generate high-quality videos that are in line with the conditions.
The term condition here can encompass different modalities and various categories. 
This paper selects three specific conditions in the experiment to verify the effectiveness of \ModelName: camera ($E$), identities ($I$), and depth ($D$). 
These conditions are selected for their substantial differences in modality representation and distribution.
Camera captures 3D scene positional changes, acting as a constraint on camera motion.
Identities, given as images, define character attributes. Depth, provided in video format, offers structural layout guidance.
We also input the text ($P$) condition with cross-attention to control the overall generation content.
Thus, the overall goal of our generation model $u_{\Theta}$ is to learn conditional distribution $p(\mathbf{x}|P,E,I,D)$.

\subsection{Condition Tokenization}

As discussed in Section~\ref{sec:overview}, \ModelName~aims to explore how conditions with different formations can be effectively composed. So we choose camera (temporal-only), identities (spatial-only), and depth (temporal-spatial) as inputs. 
These conditions are distinct from each other in feature shape, data distribution, and controlling effect, which need to be tokenized separately.
This section gives details of tokenization.

\noindent \textbf{Camera.} The input is a sequence of camera parameters $\{E_i|i = 1\cdots N\}$, where $E_i = [R_i;T_i]\in \mathbb{R}^{3\times 4}$ for the $i$th frame, $R_i \in \mathbb{R}^{3 \times 3}$ is the object orientation, $T_i \in \mathbb{R}^{3}$ is the object translation, and $N$ is the frame number. We follow CameraCtrl~\cite{cameractrl} and CamI2V~\cite{cami2v} to apply plücker embedding to facilitate the model in correlating the camera parameters with image pixels, thereby enabling precise control for visual details. Specifically, the camera parameter $E_i$ can be transferred to its plücker embedding $\{p_{u,v} = (o \times d_{u,v}, d_{u,v})|u=1\cdots H, v =1\cdots W\}$ with:

\vspace{-3mm}
$$
d_{u,v} = RK^{-1}[u,v,1]^{T}+T
$$
\vspace{-4mm}

\noindent , where $H$ is frame height, $W$ is frame width, $o \in \mathbb{R}^{3}$ is the camera center, $K \in R^{3\times 3}$ is camera intrinsic parameters.

We patchify the plücker embedding $p \in \mathbb{R}^{N \times H \times W \times 6}$ with a patch size of $16$ and get camera sequence $p_{seq} \in \mathbb{R}^{L_{p_{seq}} \times 1536}$, where $L_{p_{seq}} = N \times \tfrac{H}{16}\times \tfrac{W}{16}$. $p_{seq}$ is then mapped to hidden dimension with a convolution layer.

\noindent \textbf{Identities.} \ModelName~uses a causal 3D VAE with temporal compression rate of $4$ and spatial compression rate of $8$ to encode images and videos. Identity images $I \in \mathbb{R}^{H \times W \times 3}$ are first encoded to $I_{enc} \in \mathbb{R}^{\tfrac{H}{8} \times \tfrac{W}{8} \times 8}$ with VAE, then patchify with a patch size of $2$ to get sequence $I_{seq} \in \mathbb{R}^{L_{I_{seq}} \times 32}$, where $L_{I_{seq}} = \tfrac{H}{16}\times \tfrac{W}{16}$. 
If multiple identity images are provided, the same procedure is applied to each of them. Afterward, all identity sequences $I_{seq}$ are mapped to the hidden dimension with a convolution layer. The convolution layer is initialized using the weights from the projection layer after video patchify.

\noindent \textbf{Depth.} Depth video $D \in \mathbb{R}^{N \times H \times W \times 3}$ follow the same processing procedure with noisy video $\mathbf{x}_t \in \mathbb{R}^{N \times H \times W \times 3}$. The depth video is first encoded to $D_{enc} \in \mathbb{R}^{\tfrac{N}{4} \times \tfrac{H}{8} \times \tfrac{W}{8} \times 8}$ by VAE, then patchify to $D_{seq} \in \mathbb{R}^{L_{D_{seq}} \times 32}$ with patch size $2$, where $L_{D_{seq}} = \tfrac{N}{4}\times \tfrac{H}{16}\times \tfrac{W}{16}$. Finally, $D_{seq}$ is projected to the hidden dimension with a convolutional layer. The convolution layer is initialized using the weights from the projection layer after video patchify.

After tokenization of each condition, the sequence of noisy video $\mathbf{x}_{seq}$, camera $p_{seq}$, identities $I_{seq}$, and depth video $D_{seq}$ are concatenated along the sequential dimension, allowing joint modeling of multiple conditions.

\noindent \textbf{Discussion.} While this paper implements only three conditions, the architecture of \ModelName~is designed to easily incorporate other modalities or conditions without major structural changes. 
For example, segmentation videos and sketch videos, which exhibit representational similarities to depth videos, can employ identical tokenization techniques as used for depth.
Other modalities, such as audio, can also be tokenized into sequential representations and jointly learned with full attention mechanisms.

\subsection{Training Strategy}

\noindent \textbf{Dataset Construction.} The training of \ModelName~requires video annotation of text, camera, identities, and depth. However, since obtaining all conditions for every video is challenging, we adopt a selective annotation strategy, prioritizing label types that are most compatible with corresponding video data. 
For text labeling, we follow MiraData~\cite{miradata} and annotate text prompts using structured captions, which can include more detailed information for the video. 
For camera data, we primarily rely on ground-truth annotations, as existing automated annotation pipelines are unable to achieve sufficiently high quality. Consistent with prior research, we employ the static scene camera dataset RealEstate10K~\cite{realestate10k} for training. We observed that using only static scene camera datasets can lead to reduced human and object movement in the generated videos. To mitigate this, we further conduct quality fine-tuning using internal camera datasets that incorporate dynamic movements.
For identity annotation, we follow the data creation pipeline of ConceptMaster~\cite{conceptmaster} that includes fast elimination of unsuitable videos and fine-grained identity information extraction.
For depth annotation, we use Depth Anything~\cite{depthanything}. Finally, we use around $1$ million high-quality samples for training.

\begin{figure}[htbp]
    \centering
    \includegraphics[width=0.99\linewidth]{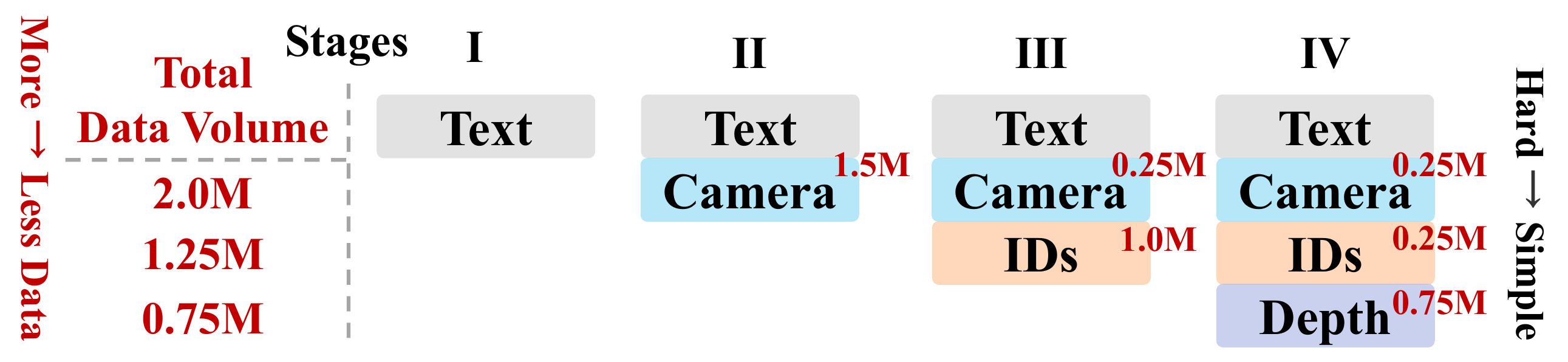}
    \vspace{-3mm}
    \caption{\textbf{Illustration of the condition training order.} We use \textcolor{brightmaroon}{red} to indicate the training data volume. M is for million.
    } 
    \vspace{-6mm}
\label{fig:training_order}
\end{figure}

\noindent \textbf{Condition Training Order.} During the pre-training phase, we noted that more challenging tasks demand extended training time and should be introduced earlier in the learning process. These challenging tasks involve complex data distributions that differ significantly from the output video, requiring the model to possess sufficient capacity to accurately capture and represent them. Conversely, introducing easier tasks too early may lead the model to prioritize learning them first, since they provide more immediate optimization feedback, which hinder the convergence of more challenging tasks. 
To address this, we adopt a progressive training strategy as shown in Figure~\ref{fig:training_order}, where we introduce difficult tasks early to ensure the model learns robust representations. Once these challenging tasks are well-trained, the easier tasks can leverage the acquired knowledge, benefiting from improved feature representations and converging more efficiently. Following this principle, we structure our training order as follows: text, camera, identities, and depth, with easier tasks using less training data volume.
After pre-training, we further refine the model through a quality fine-tuning phase to enhance motion dynamics, fine-grained controllability, and overall visual quality.

\section{Experiments}

\subsection{Evaluation Benchmark and Metrics}
\noindent \textbf{Benchmark.} To evaluate \ModelName~in multi-task video generation, we construct \BenchmarkName~with $1,400$ high-quality test cases. It comprises seven categories, each covering different condition combinations with $200$ test cases:

\noindent \textit{(1) Camera-to-Video.} We follow previous works~\cite{cameractrl,cami2v} to randomly select $200$ cases from RealEstate10k~\cite{realestate10k} test set.

\begin{figure}[htbp]
\vspace{-2mm}
    \centering
    \includegraphics[width=0.99\linewidth]{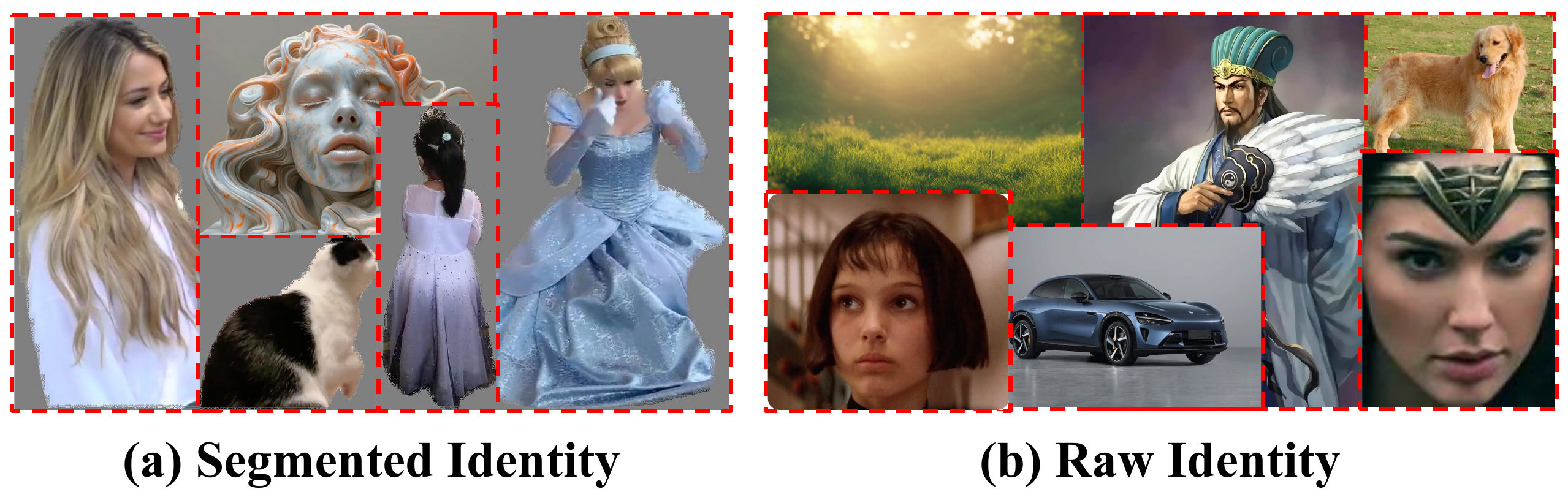}
    \vspace{-2mm}
    \caption{\textbf{Examples of two types of identity images.}
    } 
    \vspace{-5mm}
\label{fig:ids}
\end{figure}

\noindent \textit{(2) Identities-to-Video.} We collect an identities-to-video test set with two types of data. The first category uses segmented identity images (shown in Figure~\ref{fig:ids} (a)), and the second category incorporates raw images with a main identity (shown in Figure~\ref{fig:ids} (b)). Incorporating both types of test samples ensures coverage of in-domain and out-of-domain cases, leading to a more accurate model evaluation.

\begin{table*}[htbp]
\centering
\setlength\tabcolsep{3pt}
\scalebox{0.86}{
    \begin{threeparttable}
    \begin{tabular}{l|c|ccc|cc|c|ccc}
    \toprule
    \toprule
    \textbf{Metrics} & \textbf{Text}       & \multicolumn{3}{c|}{\textbf{Camera}} & \multicolumn{2}{c|}{\textbf{Identities}} & \textbf{Depth} & \multicolumn{3}{c}{\textbf{Overall Quality}}        \\ \midrule
    \textbf{Model}   & \textbf{Clip Score$\uparrow$} & \textbf{RotErr$\downarrow$}  & \textbf{TransErr$\downarrow$}  & \textbf{CamMC$\downarrow$} & \textbf{DINO-I$\uparrow$}     & \textbf{CLIP-I$\uparrow$}     & \textbf{MAE$\downarrow$}   & \textbf{Smoothness$\uparrow$} & \textbf{Dynamic$\uparrow$} & \textbf{Aesthetic$\uparrow$}  \\ \midrule
   \multicolumn{11}{c}{\textbf{Camera to Video}}          \\ \midrule
    \textbf{MotionCtrl}~\cite{motionctrl}  & 22.27   &   1.49   &   4.41     &    4.84          &             -       &   -    &    -    &  96.16    &     11.43 & 4.71            \\ 
    \textbf{CameraCtrl}~\cite{cameractrl}    &     21.36      &  1.57       &     3.88      &   4.77    &       -     &         -   &   -    &    95.16   &     13.72     &     4.66            \\ 
    \textbf{CamI2V$^\ddag$}~\cite{cami2v}     &   -     &           1.43      &   3.81    &     4.62       &      -      &   -    &   -     &    94.50    &   19.40    &  -     \\ 
    \midrule
    \textbf{\ModelName}     &    \textbf{22.97}    &     \textbf{1.20}    &   \textbf{3.31}        &   \textbf{3.98}    &      -      &       -     &    -   &  \textbf{96.40}     &     \textbf{30.53}     &  \textbf{4.95}          \\ 
    \midrule
   \multicolumn{11}{c}{\textbf{Identities to Video}}          \\ \midrule
    \textbf{ConceptMaster}~\cite{conceptmaster}        &    18.54     &       -    &   -    &      -      &      39.97      &  65.63   &    -    &   \textbf{95.05}     &    10.14    &  5.21       \\ 
    \midrule
    \textbf{\ModelName}     &    \textbf{18.64}     &    -     &       -    &    -   &      \textbf{46.22}      &     \textbf{68.59}      &   -    &  94.95  &    \textbf{16.68} & \textbf{5.46}                  \\ 
    \midrule
   \multicolumn{11}{c}{\textbf{Depth to Video}}          \\ \midrule
    \textbf{Ctrl-Adapter$^\ddag$}~\cite{ctrladapter}        &  -       &        -   &    -   &       -     &       -     &     -  &   25.63     &    94.23    & 15.47 & -              \\ 
    \textbf{ControlVideo}~\cite{controlvideo}        &     23.38    &      -     &    -   &       -     &         -   &   -    & 30.10 &     94.44   &    18.62       &    \textbf{5.91}       \\ 
    \midrule
    \textbf{\ModelName}     &    \textbf{23.40}     &   -      &      -     &    -   &      -      &     -       & \textbf{14.71}    &    \textbf{95.42}    &   \textbf{23.12}        &     5.26        \\
    \bottomrule
    \bottomrule
    \end{tabular}
    \begin{tablenotes} 
        \footnotesize 
        \item[$\ddag$] Since this method only supports image-to-video generation, we input the ground-truth first video frame into the model. Thus, frame quality metrics are not reported.
    \end{tablenotes}
    \end{threeparttable}
}
\vspace{-2mm}
\caption{\textbf{Quantitative comparison of single task video generation.} We compare \ModelName~with MotionCtrl~\cite{motionctrl}, CameraCtrl~\cite{cameractrl}, and CamI2V~\cite{cameractrl} on camera-to-video generation. For identity-to-video, due to a lack of open-source multiple identities video generation method, we compare with ConceptMaster~\cite{conceptmaster} model with the size of $1B$. We compare \ModelName~with Ctrl-Adapter~\cite{ctrladapter} and ControlVideo~\cite{controlvideo} for depth-to-video. We follow the default setting of each model for evaluation. Since most of the previous methods can generate only $16$ frames of video, we uniformly sample $16$ frames from methods that generate more than $16$ frames for comparison.}
\label{tab:single_condition_comparison}
\vspace{-3mm}
\end{table*}

\noindent \textit{(3) Depth-to-Video.} From Panda-70M~\cite{panda70m}, we randomly selected $200$ high-quality videos with significant depth variations, ensuring they were not part of the training set, and annotated their depth using Depth Anything~\cite{depthanything}.

\noindent \textit{(4) [Camera+Identities]-to-Video.} We select $200$ raw identity image pairs (Figure~\ref{fig:ids} (b)) and $200$ 3D camera trajectories from RealEstate10k~\cite{realestate10k} test set. These identity images and camera trajectories differ from those in (1) and (2).

\noindent \textit{(5) [Camera+Depth]-to-Video.} We randomly select $200$ cases from RealEstate10k~\cite{realestate10k} test set and annotate depth with Depth  Anything~\cite{depthanything}. Note that these camera trajectories differ from those in (1) and (4) to increase test diversity.

\noindent \textit{(6) [Identities+Depth]-to-Video.} We collect identity-video pairs following (2) and annotate with Depth Anything~\cite{depthanything}, with identity images differing from those in (2) and (4).

\noindent \textit{(7) [Camera+Identities+Depth]-to-Video.} We first collect identity-depth-video pairs following (6), then annotate camera parameters with GLOMAP~\cite{glomap}. The samples we selected are different from those in (6) to enhance diversity.

\noindent \textbf{Metrics.} We employ $10$ metrics across five key aspects: text alignment, camera control, identity similarity, depth control, and overall video quality. Following prior work~\cite{conceptmaster}, we use CLIP similarity~\cite{clip} for text alignment. For camera control, we adopt RotErr, TransErr, and CamMC, as in CamI2V~\cite{cami2v}. Identity similarity~\cite{dreambooth} is assessed using DINO-I~\cite{dino} and CLIP-I~\cite{clip}. Depth control is measured via Mean Absolute Error (MAE), following previous works~\cite{sparsectrl,controlvideo}. We incorporate three metrics from MiraData~\cite{miradata} to evaluate video quality: frame CLIP similarity~\cite{clip} for smoothness, optical flow motion distance~\cite{raft} for dynamics, and the LAION-Aesthetic~\cite{laion} model for aesthetic assessment. Details are in the appendix.

\begin{figure*}[htbp]
    \centering
    \includegraphics[width=0.96\linewidth]{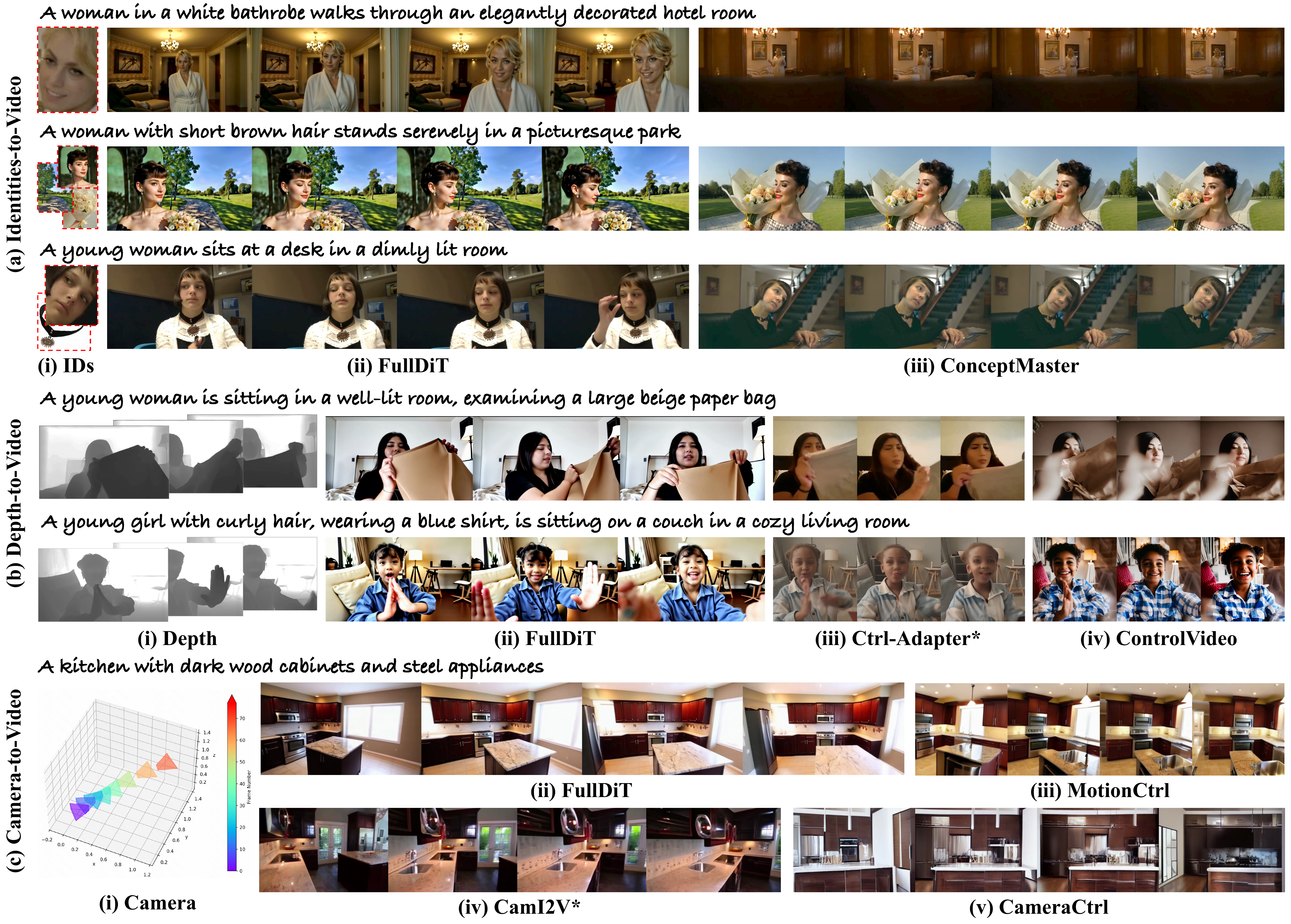}
    \vspace{-4mm}
    \caption{\textbf{Qualitative comparison of \ModelName~and previous single control video generation methods.} We present identity-to-video results compared with ConceptMaster~\cite{conceptmaster}, depth-to-video results compared with Ctrl-Adapter~\cite{ctrladapter} and ControlVideo~\cite{controlvideo}, and camera-to-video results compared with MotionCtrl~\cite{motionctrl}, CamI2V~\cite{cami2v}, and CameraCtrl~\cite{cameractrl}. Results denoted with * are image-to-video methods.
    } 
    \vspace{-6mm}
\label{fig:single_condition}
\end{figure*}

\vspace{-1mm}
\subsection{Implementation Details}
\vspace{-1mm}

We train \ModelName~based on an internal text-to-video diffusion model with approximately $1B$ parameters. We use a small parameter size to ensure fair comparison with previous methods and facilitate reproducibility.
Since the training videos vary in size and length, we resize and pad all videos to a unified resolution in each batch and sample $77$ frames. We apply attention masks and loss masks to ensure proper training.
We employ the Adam optimizer with a learning rate of $1 \times 10^{-5}$ and train on a cluster of $64$ NVIDIA H800 GPUs. The $1B$ model requires approximately $32,000$ steps of training with $20$ frames of camera control, a maximum of $3$ identities, and $21$ frames of depth conditions. Camera and depth control are evenly sampled from $77$ frames.
The detailed training data volume is shown in Figure~\ref{fig:training_order}.
For inference of the $1B$ model, we use a resolution of $384 \times 672$ with $77$ frames (approximately $5$ seconds with FPS of $15$). We set the inference step number as $50$ and the classifier free guidance scale as $5$.

\vspace{-2mm}

\subsection{Comparison with Previous Methods}
\vspace{-2mm}

This section is to validate the superior performance of \ModelName~in comparison to previous adapter-based methods. We evaluate \ModelName~against prior single-condition guided video generation methods on the camera-to-video, identities-to-video, and depth-to-video subsets of our benchmark \BenchmarkName. Given the absence of open-source multiple conditions to video generation methods suitable for comparison, we do not provide comparisons of \ModelName~with previous methods. We put quantitative results of \ModelName~on other subsets of \BenchmarkName~in the appendix.

\noindent \textbf{Quantitative Comparison on Single-Task Generation.} For camera-to-video, we compare \ModelName~with MotionCtrl~\cite{motionctrl}, CameraCtrl~\cite{cameractrl}, and CamI2V~\cite{cami2v}. All of these models are trained on the RealEstate10k~\cite{realestate10k} dataset, ensuring a consistent and fair training data setup for camera conditions. For identities-to-video, due to the absence of an open-source multi-identity video generation model of comparable parameter scale, we benchmark against the $1B$ ConceptMaster~\cite{conceptmaster}, using identical training data with \ModelName. 
This ensures a fair comparison of the same model architecture and training data, which further validates the advantages of full attention.
For depth-to-video, We compare with Ctrl-Adapter~\cite{ctrladapter} and ControlVideo~\cite{controlvideo}.
Results show that although the \ModelName~integrates multiple conditions, it still achieves state-of-the-art performance on controlling metrics (\textit{i.e.}, text, camera, identities, and depth controls), thereby validating the effectiveness of the \ModelName.
For the overall quality metrics, \ModelName~outperforms previous methods across the majority. The smoothness of \ModelName~is slightly lower than that of ConceptMaster since the calculation of smoothness is based on CLIP similarity between adjacent frames. As \ModelName~exhibits significantly greater dynamics compared to ConceptMaster, the smoothness metric is impacted by the large variations between adjacent frames. For the aesthetic score, since the rating model favors images in painting style and ControlVideo typically generates videos in this style, it achieves a high score in aesthetics.

\noindent \textbf{Qualitative Comparison on Single-Task Generation.} As illustrated in Figure~\ref{fig:single_condition} (a), \ModelName~demonstrates superior identity preservation and generates videos with better dynamics and visual quality compared to ConceptMaster~\cite{conceptmaster}. Since ConceptMaster and \ModelName~are trained on the same backbone, this highlights the effectiveness of condition injection with full attention. We further present additional comparisons of depth-to-video and camera-to-video in Figure~\ref{fig:single_condition} (b) and (c). The results demonstrate the superior controllability and generation quality of \ModelName~compared to existing depth-to-video and camera-to-video methods.

\begin{figure*}[htbp]
    \centering
    \includegraphics[width=0.99\linewidth]{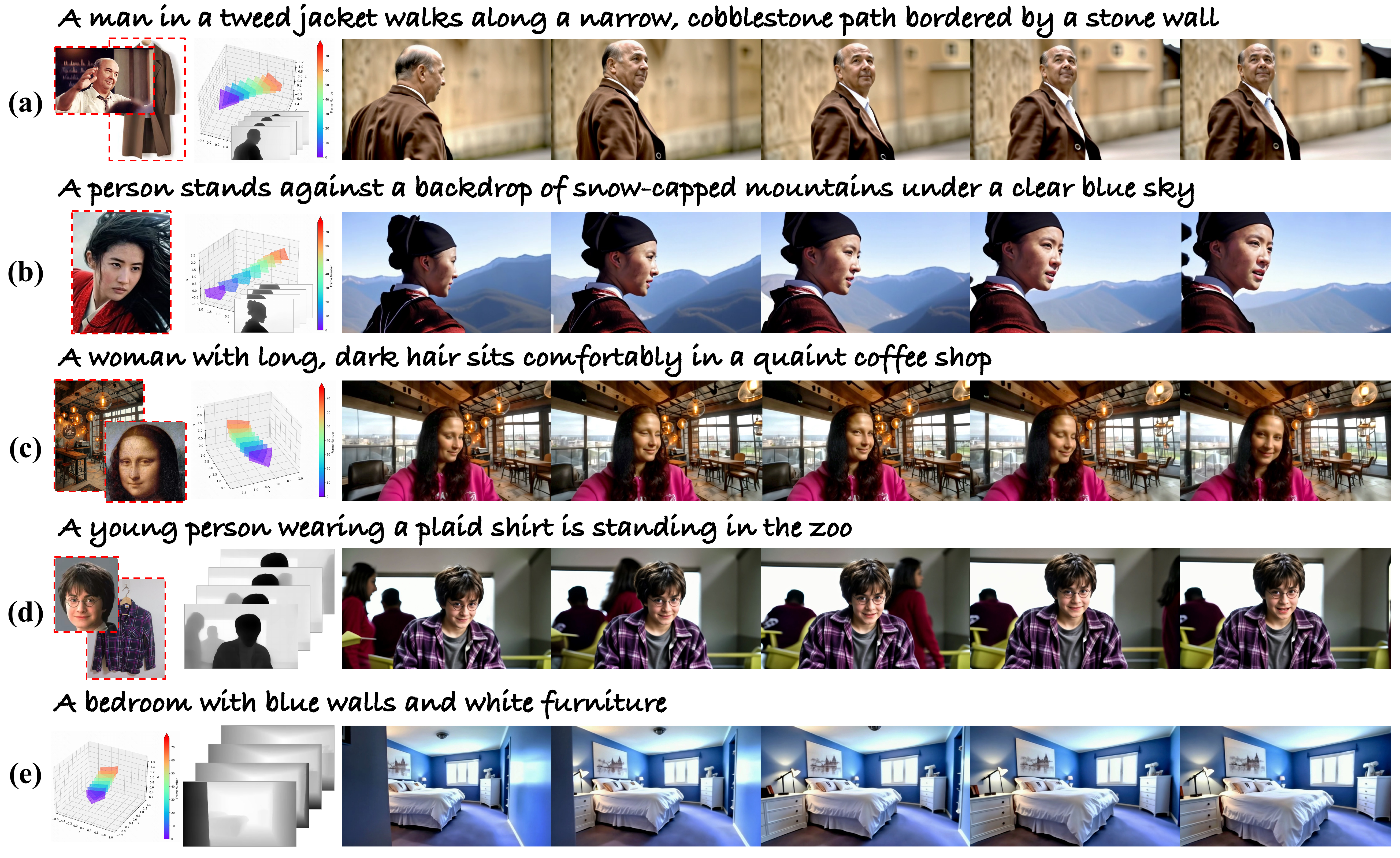}
    \vspace{-2mm}
    \caption{\textbf{Qualitative results of \ModelName~with multiple control signals.} We show camera+identity+depth-to-video in (a) and (b), camera+identity-to-video in (c), identity+depth-to-video in (d), and camera+depth-to-video in (e).
    } 
    \vspace{-4mm}
\label{fig:muliple_condition}
\end{figure*}

\subsection{Scalability and Emergent ability of FullDiT}

\begin{figure}[htbp]
    \vspace{-2mm}
    \centering
    \includegraphics[width=0.95\linewidth]{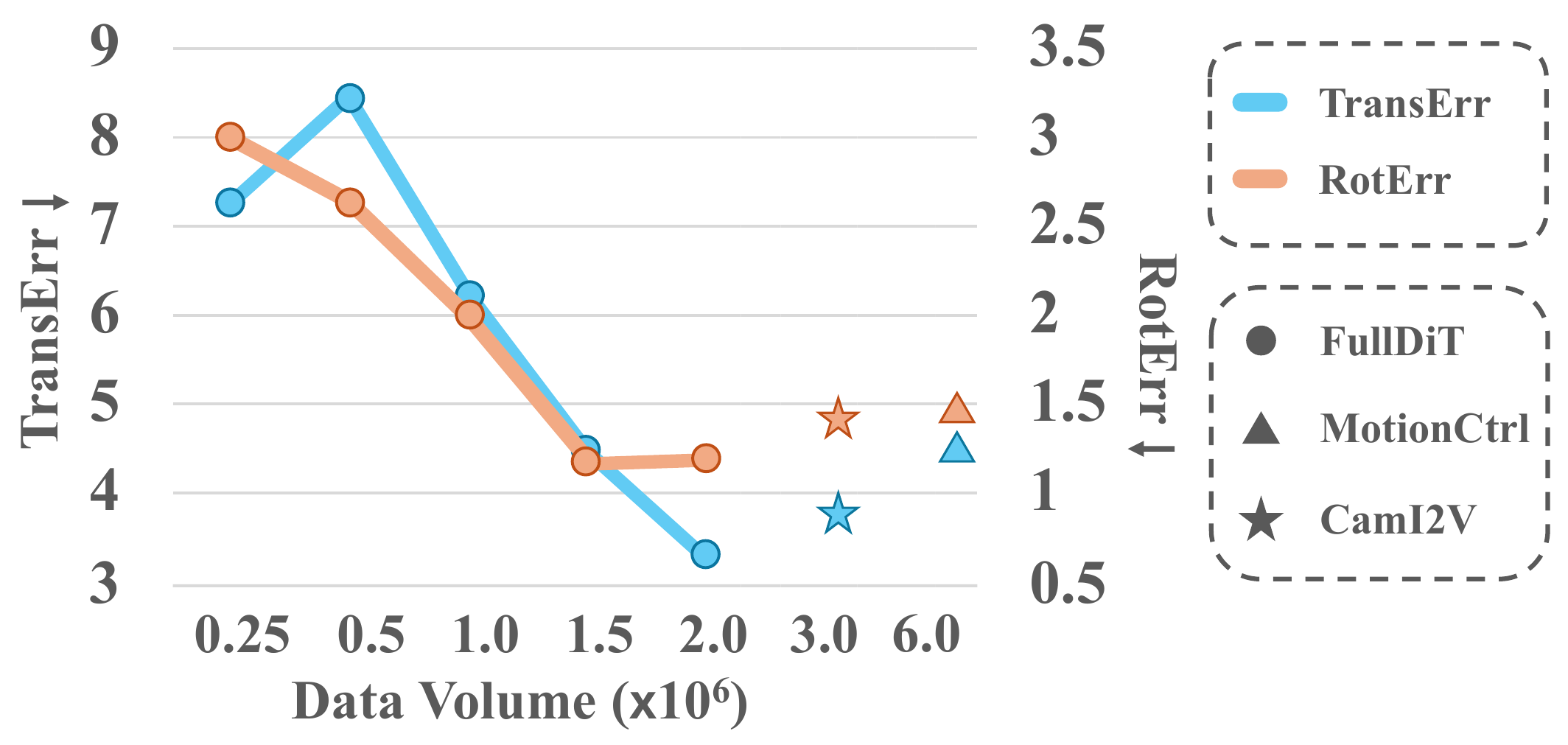}
    \vspace{-3mm}
    \caption{\textbf{Camera-to-Video Performance with the Increase of Training Data Volume.} We also show the data volume and performance of MotionCtrl~\cite{motionctrl} and CamI2V~\cite{cami2v} for comparison.
    } 
    \vspace{-4mm}
\label{fig:scalability}
\end{figure}

\noindent \textbf{Scalability.} Shown in figure~\ref{fig:scalability}, results of \ModelName~camera-to-video get better on both TransErr and RotErr as training data volume grows, which illustrates the scalability of \ModelName. In comparison, MotionCtrl~\cite{motionctrl} employed data volume of $6.4\times 10^6$ and CameI2V~\cite{cami2v} used data volume of $3.2\times 10^6$, yet both performed worse than \ModelName. This further shows the effectiveness of full attention

\noindent \textbf{Combination and Emergent Ability.} We demonstrate the results of feeding multiple conditions into \ModelName~in Figures~\ref{fig:muliple_condition} and~\ref{fig:teaser}. These results highlight \ModelName's ability to combine multiple condition inputs, even without training data that encompasses all conditions concurrently. For instance, our training data contains no videos with both camera and identity annotations. But as shown in Figure~\ref{fig:muliple_condition}{c}, \ModelName~can effectively generate videos that faithfully reflect both camera and identity inputs. This demonstrates the emergent ability of \ModelName~on unseen tasks.

\vspace{-1mm}
\subsection{Ablation Study}

\vspace{-1mm}

\begin{table}[htbp]
\centering
\setlength\tabcolsep{2pt}
\scalebox{0.66}{
    \begin{threeparttable}
    \begin{tabular}{l|ccc|cc|c}
    \toprule
    \toprule
    \textbf{Metrics}      & \multicolumn{3}{c|}{\textbf{Camera}} & \multicolumn{2}{c|}{\textbf{Identities}} & \textbf{Depth}        \\ \midrule
    \textbf{Stage}   &\textbf{RotErr$\downarrow$}  & \textbf{TransErr$\downarrow$}  & \textbf{CamMC$\downarrow$} & \textbf{DINO-I$\uparrow$}     & \textbf{CLIP-I$\uparrow$}     & \textbf{MAE$\downarrow$}    \\ \midrule
    \textbf{Depth$\rightarrow$Camera$\rightarrow$IDs}           & 2.50      &    6.57     &  8.17  &  36.46        &    64.56    &      14.76        \\ 
    \textbf{IDs$\rightarrow$Camera$\rightarrow$Depth}           &    2.46   &    7.43     &  8.52  &     42.71     &    65.99    &   14.94           \\ 
    \textbf{Camera$\rightarrow$IDs$\rightarrow$Depth}         &  \textbf{1.20}           &  \textbf{3.31}   &       \textbf{3.98}    &    \textbf{46.22}    &  \textbf{68.59}    &   \textbf{14.71}      \\ 
    \bottomrule
    \bottomrule
    \end{tabular}
    \end{threeparttable}
}
\vspace{-2mm}
\caption{\textbf{Ablation on condition training order.}}
\vspace{-1mm}
\label{tab:ablation_condition_order_single_subset}

\end{table}

\noindent\textbf{Impact of Condition Training Order.} We train three sets of models with different conditions training orders, using the same data column for each condition: (1) identities, followed by camera, then depth; (2) depth, followed by camera, then identities; and (3) camera, followed by identities, then depth. We evaluate our model on the camera-to-video, identities-to-video, and depth-to-video subset of \BenchmarkName. Results in Table~\ref{tab:ablation_condition_order_single_subset} validate our claim that more challenging tasks require additional training and should be introduced earlier. Especially, introducing the camera condition too late would significantly reduce its controllability.

\noindent\textbf{Impact of Number of Training Stages.} We further analyze the impact of using multiple stages of training, as well as the influence of later stages on earlier stages' conditions, all under the same data volume. We evaluate our model on the camera-to-video, identities-to-video, and depth-to-video subset of \BenchmarkName. Table~\ref{tab:ablation_training_stages} shows that multi-stage training leads to better condition control. 
Specifically, by comparing one-stage and two-stage training, we observe that isolating the camera as an independent training stage significantly improves camera control metrics.

\begin{table}[htbp]
\centering
\setlength\tabcolsep{2pt}
\scalebox{0.66}{
    \begin{threeparttable}
    \begin{tabular}{l|ccc|cc|c}
    \toprule
    \toprule
    \textbf{Metrics}      & \multicolumn{3}{c|}{\textbf{Camera}} & \multicolumn{2}{c|}{\textbf{Identities}} & \textbf{Depth}        \\ \midrule
    \textbf{Stage}   &\textbf{RotErr$\downarrow$}  & \textbf{TransErr$\downarrow$}  & \textbf{CamMC$\downarrow$} & \textbf{DINO-I$\uparrow$}     & \textbf{CLIP-I$\uparrow$}     & \textbf{MAE$\downarrow$}    \\ \midrule
    \textbf{I: Camera+ID+Depth}       &     2.69   &    6.19     &  8.21 &     35.42     &   59.49    &   32.88      \\ \midrule
     \textbf{I: Camera}   &  \textbf{1.19}  &  4.49    &      5.01      &    -    &   -    &   -    \\ 
    \textbf{II: Camera+ID+Depth}           &   1.23    &     4.14     &  4.78  &    37.21      &   65.85       &   15.81            \\ \midrule
     \textbf{I: Camera}   &   \textbf{1.19}  &  4.49    &      5.01      &    -    &   -    &   -      \\ 
    \textbf{II: Camera+ID}        &   1.23    &    4.20      &  4.82  &    42.83      &   64.99       &     -     \\ 
    \textbf{III: Camera+ID+Depth}       &  1.20           &  \textbf{3.31}   &       \textbf{3.98}    &    \textbf{46.22}     &  \textbf{68.59}    &   \textbf{14.71}      \\ 
    \bottomrule
    \bottomrule
    \end{tabular}
    \end{threeparttable}
}
\vspace{-3mm}
\caption{\textbf{Ablation on number of training stages.}}
\label{tab:ablation_training_stages}
\end{table}

\noindent \textbf{Impact of Model Architecture.} To fairly compare the performance between adapter-based approaches and \ModelName~within the same architecture and training data, we followed CameraCtrl~\cite{cameractrl} to implement a camera-to-video model on our model architecture. This model starts with the same text-to-video weights as \ModelName~and uses only camera data for training. Table~\ref{tab:fulldit_vs_adapter} shows that, although \ModelName~is trained with three conditions, it still outperforms the adapter architecture in terms of camera control.

\begin{table}[htbp]
\centering
\setlength\tabcolsep{2pt}
\scalebox{0.6}{
    \begin{threeparttable}
    \begin{tabular}{l|c|ccc|ccc}
    \toprule
    \toprule
    \textbf{Metrics}     & \textbf{Text} & \multicolumn{3}{c|}{\textbf{Camera}} & \multicolumn{3}{c}{\textbf{Overall Quality}}     \\ \midrule
    \textbf{Model}   & \textbf{Clip Score}$\uparrow$ & \textbf{RotErr}$\downarrow$  & \textbf{TransErr}$\downarrow$  & \textbf{CamMC}$\downarrow$ & 
    \textbf{Smoothness}$\uparrow$    & \textbf{Dynamic}$\uparrow$    & \textbf{Aesthetic}$\uparrow$  \\ \midrule
    \textbf{Adapter}          & 22.58  &  1.28       &   3.35    &      4.17      &   \textbf{96.41}         &  28.42     &    4.88       \\ 
    \midrule
    \textbf{\ModelName}         & \textbf{22.97} &  \textbf{1.20}      &    \textbf{3.31}       &   \textbf{3.98}    &     96.40       &  \textbf{30.53}           &   \textbf{4.95}      \\ 
    \bottomrule
    \bottomrule
    \end{tabular}
    \end{threeparttable}
}
\vspace{-2mm}
\caption{\textbf{Comparing \ModelName~with adapter-based architecture.}}
\label{tab:fulldit_vs_adapter}
\vspace{-5mm}
\end{table}

\section{Conclusion}

We introduced \ModelName, a novel video generative foundation model leveraging unified full-attention to seamlessly integrate multimodal conditions. \ModelName~resolves adapter-based limitations such as branch conflicts and parameter redundancy, enabling scalable multi-task and multimodal control. We also provided \BenchmarkName, the first comprehensive benchmark for evaluating multi-condition video generation. Extensive experiments demonstrated \ModelName’s state-of-the-art performance and emergent capabilities.

\noindent \textbf{Limitations and Future Works.} Despite the strong performance, there are some limitations that need further study: 

(1) In this work, we only explore control conditions of the camera, identities, and depth information. We did not further investigate other conditions and modalities such as audio, speech, point cloud, object bounding boxes, optical flow, etc. Although the design of \ModelName~can seamlessly integrate other modalities with minimal architecture modification, how to quickly and cost-effectively adapt existing models to new conditions and modalities is still an important question that warrants further exploration.

(2) The design philosophy of \ModelName~is inherited from MMDiT~\cite{sd3}, which utilizes self-attention to process text and images simultaneously. Compared to MMDiT, \ModelName~takes a further step in exploring a more unified model architecture and a more scalable design for multiple control inputs. However, due to the structural constraints of the pre-trained model, we incorporate text through cross-attention, and \ModelName~is not adapted directly from the MMDiT architecture. We anticipate future work to explore a more flexible integration of MMDiT and \ModelName~architectures.

{
    \small
    \bibliographystyle{ieeenat_fullname}
    \bibliography{main}

\begin{thebibliography}{71}
\providecommand{\natexlab}[1]{#1}
\providecommand{\url}[1]{\texttt{#1}}
\expandafter\ifx\csname urlstyle\endcsname\relax
  \providecommand{\doi}[1]{doi: #1}\else
  \providecommand{\doi}{doi: \begingroup \urlstyle{rm}\Url}\fi

\bibitem[cav(2024)]{cavia}
Cavia: Camera-controllable multi-view video diffusion with view-integrated attention.
\newblock \emph{arXiv preprint arXiv:2410}, 2024.

\bibitem[Bahmani et~al.(2024)Bahmani, Skorokhodov, Siarohin, Menapace, Qian, Vasilkovsky, Lee, Wang, Zou, Tagliasacchi, et~al.]{vd3d}
Sherwin Bahmani, Ivan Skorokhodov, Aliaksandr Siarohin, Willi Menapace, Guocheng Qian, Michael Vasilkovsky, Hsin-Ying Lee, Chaoyang Wang, Jiaxu Zou, Andrea Tagliasacchi, et~al.
\newblock Vd3d: Taming large video diffusion transformers for 3d camera control.
\newblock \emph{arXiv preprint arXiv:2407.12781}, 2024.

\bibitem[Blattmann et~al.(2023)Blattmann, Rombach, Ling, Dockhorn, Kim, Fidler, and Kreis]{alignyourlatents}
Andreas Blattmann, Robin Rombach, Huan Ling, Tim Dockhorn, Seung~Wook Kim, Sanja Fidler, and Karsten Kreis.
\newblock Align your latents: High-resolution video synthesis with latent diffusion models.
\newblock In \emph{Proceedings of the IEEE/CVF Conference on Computer Vision and Pattern Recognition}, pages 22563--22575, 2023.

\bibitem[Brooks et~al.(2024)Brooks, Peebles, Holmes, DePue, Guo, Jing, Schnurr, Taylor, Luhman, Luhman, Ng, Wang, and Ramesh]{sora}
Tim Brooks, Bill Peebles, Connor Holmes, Will DePue, Yufei Guo, Li Jing, David Schnurr, Joe Taylor, Troy Luhman, Eric Luhman, Clarence Ng, Ricky Wang, and Aditya Ramesh.
\newblock Video generation models as world simulators, 2024.

\bibitem[Caron et~al.(2021)Caron, Touvron, Misra, J{\'e}gou, Mairal, Bojanowski, and Joulin]{dino}
Mathilde Caron, Hugo Touvron, Ishan Misra, Herv{\'e} J{\'e}gou, Julien Mairal, Piotr Bojanowski, and Armand Joulin.
\newblock Emerging properties in self-supervised vision transformers.
\newblock In \emph{Proceedings of the IEEE/CVF international conference on computer vision}, pages 9650--9660, 2021.

\bibitem[Chen et~al.(2024{\natexlab{a}})Chen, Zhang, Cun, Xia, Wang, Weng, and Shan]{videocrafter2}
Haoxin Chen, Yong Zhang, Xiaodong Cun, Menghan Xia, Xintao Wang, Chao Weng, and Ying Shan.
\newblock Videocrafter2: Overcoming data limitations for high-quality video diffusion models, 2024{\natexlab{a}}.

\bibitem[Chen et~al.(2024{\natexlab{b}})Chen, Siarohin, Menapace, Deyneka, Chao, Jeon, Fang, Lee, Ren, Yang, et~al.]{panda70m}
Tsai-Shien Chen, Aliaksandr Siarohin, Willi Menapace, Ekaterina Deyneka, Hsiang-wei Chao, Byung~Eun Jeon, Yuwei Fang, Hsin-Ying Lee, Jian Ren, Ming-Hsuan Yang, et~al.
\newblock Panda-70m: Captioning 70m videos with multiple cross-modality teachers.
\newblock In \emph{Proceedings of the IEEE/CVF Conference on Computer Vision and Pattern Recognition}, pages 13320--13331, 2024{\natexlab{b}}.

\bibitem[Chen et~al.(2025)Chen, Siarohin, Menapace, Fang, Lee, Skorokhodov, Aberman, Zhu, Yang, and Tulyakov]{VideoAlchemist}
Tsai-Shien Chen, Aliaksandr Siarohin, Willi Menapace, Yuwei Fang, Kwot~Sin Lee, Ivan Skorokhodov, Kfir Aberman, Jun-Yan Zhu, Ming-Hsuan Yang, and Sergey Tulyakov.
\newblock Multi-subject open-set personalization in video generation.
\newblock \emph{arXiv preprint arXiv:2501.06187}, 2025.

\bibitem[Chen et~al.(2024{\natexlab{c}})Chen, Zhang, Zhang, Zhou, Kim, Liu, Li, Zhang, Zhao, Wang, et~al.]{unireal}
Xi Chen, Zhifei Zhang, He Zhang, Yuqian Zhou, Soo~Ye Kim, Qing Liu, Yijun Li, Jianming Zhang, Nanxuan Zhao, Yilin Wang, et~al.
\newblock Unireal: Universal image generation and editing via learning real-world dynamics.
\newblock \emph{arXiv preprint arXiv:2412.07774}, 2024{\natexlab{c}}.

\bibitem[Dai et~al.(2023)Dai, Hou, Ma, Tsai, Wang, Wang, Zhang, Vandenhende, Wang, Dubey, et~al.]{emu}
Xiaoliang Dai, Ji Hou, Chih-Yao Ma, Sam Tsai, Jialiang Wang, Rui Wang, Peizhao Zhang, Simon Vandenhende, Xiaofang Wang, Abhimanyu Dubey, et~al.
\newblock Emu: Enhancing image generation models using photogenic needles in a haystack.
\newblock \emph{arXiv preprint arXiv:2309.15807}, 2023.

\bibitem[Ding et~al.(2022)Ding, Qin, Yang, Wei, Yang, Su, Hu, Chen, Chan, Chen, et~al.]{deltatuning}
Ning Ding, Yujia Qin, Guang Yang, Fuchao Wei, Zonghan Yang, Yusheng Su, Shengding Hu, Yulin Chen, Chi-Min Chan, Weize Chen, et~al.
\newblock Delta tuning: A comprehensive study of parameter efficient methods for pre-trained language models.
\newblock \emph{arXiv preprint arXiv:2203.06904}, 2022.

\bibitem[Esser et~al.(2024)Esser, Kulal, Blattmann, Entezari, M{\"u}ller, Saini, Levi, Lorenz, Sauer, Boesel, et~al.]{sd3}
Patrick Esser, Sumith Kulal, Andreas Blattmann, Rahim Entezari, Jonas M{\"u}ller, Harry Saini, Yam Levi, Dominik Lorenz, Axel Sauer, Frederic Boesel, et~al.
\newblock Scaling rectified flow transformers for high-resolution image synthesis.
\newblock In \emph{Forty-first International Conference on Machine Learning}, 2024.

\bibitem[Ge et~al.(2024)Ge, Zhao, Zhu, Ge, Yi, Song, Li, Ding, and Shan]{seedx}
Yuying Ge, Sijie Zhao, Jinguo Zhu, Yixiao Ge, Kun Yi, Lin Song, Chen Li, Xiaohan Ding, and Ying Shan.
\newblock Seed-x: Multimodal models with unified multi-granularity comprehension and generation.
\newblock \emph{arXiv preprint arXiv:2404.14396}, 2024.

\bibitem[Geng et~al.(2024)Geng, Herrmann, Hur, Cole, Zhang, Pfaff, Lopez-Guevara, Doersch, Aytar, Rubinstein, et~al.]{motionprompting}
Daniel Geng, Charles Herrmann, Junhwa Hur, Forrester Cole, Serena Zhang, Tobias Pfaff, Tatiana Lopez-Guevara, Carl Doersch, Yusuf Aytar, Michael Rubinstein, et~al.
\newblock Motion prompting: Controlling video generation with motion trajectories.
\newblock \emph{arXiv preprint arXiv:2412.02700}, 2024.

\bibitem[Girdhar et~al.(2023)Girdhar, Singh, Brown, Duval, Azadi, Rambhatla, Shah, Yin, Parikh, and Misra]{emuvideo}
Rohit Girdhar, Mannat Singh, Andrew Brown, Quentin Duval, Samaneh Azadi, Sai~Saketh Rambhatla, Akbar Shah, Xi Yin, Devi Parikh, and Ishan Misra.
\newblock Emu video: Factorizing text-to-video generation by explicit image conditioning.
\newblock \emph{arXiv preprint arXiv:2311.10709}, 2023.

\bibitem[Guo et~al.(2025)Guo, Yang, Rao, Agrawala, Lin, and Dai]{sparsectrl}
Yuwei Guo, Ceyuan Yang, Anyi Rao, Maneesh Agrawala, Dahua Lin, and Bo Dai.
\newblock Sparsectrl: Adding sparse controls to text-to-video diffusion models.
\newblock In \emph{European Conference on Computer Vision}, pages 330--348. Springer, 2025.

\bibitem[Gupta et~al.(2023)Gupta, Yu, Sohn, Gu, Hahn, Fei-Fei, Essa, Jiang, and Lezama]{walt}
Agrim Gupta, Lijun Yu, Kihyuk Sohn, Xiuye Gu, Meera Hahn, Li Fei-Fei, Irfan Essa, Lu Jiang, and Jos{\'e} Lezama.
\newblock Photorealistic video generation with diffusion models.
\newblock \emph{arXiv preprint arXiv:2312.06662}, 2023.

\bibitem[He et~al.(2024{\natexlab{a}})He, Xu, Guo, Wetzstein, Dai, Li, and Yang]{cameractrl}
Hao He, Yinghao Xu, Yuwei Guo, Gordon Wetzstein, Bo Dai, Hongsheng Li, and Ceyuan Yang.
\newblock Cameractrl: Enabling camera control for text-to-video generation.
\newblock \emph{arXiv preprint arXiv:2404.02101}, 2024{\natexlab{a}}.

\bibitem[He et~al.(2024{\natexlab{b}})He, Liu, Qian, Wang, Hu, Cao, Yan, and Zhang]{idanimator}
Xuanhua He, Quande Liu, Shengju Qian, Xin Wang, Tao Hu, Ke Cao, Keyu Yan, and Jie Zhang.
\newblock Id-animator: Zero-shot identity-preserving human video generation.
\newblock \emph{arXiv preprint arXiv:2404.15275}, 2024{\natexlab{b}}.

\bibitem[Hu(2024)]{animateanyone}
Li Hu.
\newblock Animate anyone: Consistent and controllable image-to-video synthesis for character animation.
\newblock In \emph{Proceedings of the IEEE/CVF Conference on Computer Vision and Pattern Recognition}, pages 8153--8163, 2024.

\bibitem[Hu and Xu(2023)]{videocontrolnet}
Zhihao Hu and Dong Xu.
\newblock Videocontrolnet: A motion-guided video-to-video translation framework by using diffusion model with controlnet.
\newblock \emph{arXiv preprint arXiv:2307.14073}, 2023.

\bibitem[Huang et~al.(2024)Huang, Wang, Wu, Dou, Shi, Feng, Liang, Liu, and Zhou]{groupdiffusion}
Lianghua Huang, Wei Wang, Zhi-Fan Wu, Huanzhang Dou, Yupeng Shi, Yutong Feng, Chen Liang, Yu Liu, and Jingren Zhou.
\newblock Group diffusion transformers are unsupervised multitask learners.
\newblock 2024.

\bibitem[Huang et~al.(2025)Huang, Yuan, Liu, Wang, Wang, Zhang, Wan, Zhang, and Gai]{conceptmaster}
Yuzhou Huang, Ziyang Yuan, Quande Liu, Qiulin Wang, Xintao Wang, Ruimao Zhang, Pengfei Wan, Di Zhang, and Kun Gai.
\newblock Conceptmaster: Multi-concept video customization on diffusion transformer models without test-time tuning.
\newblock \emph{arXiv preprint arXiv:2501.04698}, 2025.

\bibitem[Ju et~al.(2023)Ju, Zeng, Zhao, Wang, Zhang, and Xu]{humansd}
Xuan Ju, Ailing Zeng, Chenchen Zhao, Jianan Wang, Lei Zhang, and Qiang Xu.
\newblock Humansd: A native skeleton-guided diffusion model for human image generation.
\newblock In \emph{Proceedings of the IEEE/CVF International Conference on Computer Vision}, pages 15988--15998, 2023.

\bibitem[Ju et~al.(2024)Ju, Gao, Zhang, Yuan, Wang, Zeng, Xiong, Xu, and Shan]{miradata}
Xuan Ju, Yiming Gao, Zhaoyang Zhang, Ziyang Yuan, Xintao Wang, Ailing Zeng, Yu Xiong, Qiang Xu, and Ying Shan.
\newblock Miradata: A large-scale video dataset with long durations and structured captions.
\newblock \emph{arXiv preprint arXiv:2407.06358}, 2024.

\bibitem[Kondratyuk et~al.(2023)Kondratyuk, Yu, Gu, Lezama, Huang, Schindler, Hornung, Birodkar, Yan, Chiu, et~al.]{videopoet}
Dan Kondratyuk, Lijun Yu, Xiuye Gu, Jos{\'e} Lezama, Jonathan Huang, Grant Schindler, Rachel Hornung, Vighnesh Birodkar, Jimmy Yan, Ming-Chang Chiu, et~al.
\newblock Videopoet: A large language model for zero-shot video generation.
\newblock \emph{arXiv preprint arXiv:2312.14125}, 2023.

\bibitem[Kong et~al.(2024)Kong, Tian, Zhang, Min, Dai, Zhou, Xiong, Li, Wu, Zhang, et~al.]{hunyuanvideo}
Weijie Kong, Qi Tian, Zijian Zhang, Rox Min, Zuozhuo Dai, Jin Zhou, Jiangfeng Xiong, Xin Li, Bo Wu, Jianwei Zhang, et~al.
\newblock Hunyuanvideo: A systematic framework for large video generative models.
\newblock \emph{arXiv preprint arXiv:2412.03603}, 2024.

\bibitem[Le et~al.(2024)Le, Pham, Lee, Clark, Kembhavi, Mandt, Krishna, and Lu]{onediffusion}
Duong~H. Le, Tuan Pham, Sangho Lee, Christopher Clark, Aniruddha Kembhavi, Stephan Mandt, Ranjay Krishna, and Jiasen Lu.
\newblock One diffusion to generate them all, 2024.

\bibitem[Lei et~al.(2024)Lei, Wang, Li, Zhang, Wang, and Xu]{animateanything}
Guojun Lei, Chi Wang, Hong Li, Rong Zhang, Yikai Wang, and Weiwei Xu.
\newblock Animateanything: Consistent and controllable animation for video generation.
\newblock \emph{arXiv preprint arXiv:2411.10836}, 2024.

\bibitem[Li et~al.(2024)Li, Kallidromitis, Gokul, Liao, Kato, Kozuka, and Grover]{omniflow}
Shufan Li, Konstantinos Kallidromitis, Akash Gokul, Zichun Liao, Yusuke Kato, Kazuki Kozuka, and Aditya Grover.
\newblock Omniflow: Any-to-any generation with multi-modal rectified flows.
\newblock \emph{arXiv preprint arXiv:2412.01169}, 2024.

\bibitem[Lin et~al.(2024{\natexlab{a}})Lin, Ge, Cheng, Li, Zhu, Wang, He, Ye, Yuan, Chen, et~al.]{opensoraplan}
Bin Lin, Yunyang Ge, Xinhua Cheng, Zongjian Li, Bin Zhu, Shaodong Wang, Xianyi He, Yang Ye, Shenghai Yuan, Liuhan Chen, et~al.
\newblock Open-sora plan: Open-source large video generation model.
\newblock \emph{arXiv preprint arXiv:2412.00131}, 2024{\natexlab{a}}.

\bibitem[Lin et~al.(2025)Lin, Jiang, Yang, Zheng, and Liang]{omnihuman}
Gaojie Lin, Jianwen Jiang, Jiaqi Yang, Zerong Zheng, and Chao Liang.
\newblock Omnihuman-1: Rethinking the scaling-up of one-stage conditioned human animation models.
\newblock \emph{arXiv preprint arXiv:2502.01061}, 2025.

\bibitem[Lin et~al.(2024{\natexlab{b}})Lin, Cho, Zala, and Bansal]{ctrladapter}
Han Lin, Jaemin Cho, Abhay Zala, and Mohit Bansal.
\newblock Ctrl-adapter: An efficient and versatile framework for adapting diverse controls to any diffusion model.
\newblock \emph{arXiv preprint arXiv:2404.09967}, 2024{\natexlab{b}}.

\bibitem[Ling et~al.(2024)Ling, Bu, Zhang, Dong, Zang, Wu, Chen, Wang, and Jin]{motionclone}
Pengyang Ling, Jiazi Bu, Pan Zhang, Xiaoyi Dong, Yuhang Zang, Tong Wu, Huaian Chen, Jiaqi Wang, and Yi Jin.
\newblock Motionclone: Training-free motion cloning for controllable video generation.
\newblock \emph{arXiv preprint arXiv:2406.05338}, 2024.

\bibitem[Lipman et~al.(2022)Lipman, Chen, Ben-Hamu, Nickel, and Le]{flowmatching}
Yaron Lipman, Ricky~TQ Chen, Heli Ben-Hamu, Maximilian Nickel, and Matt Le.
\newblock Flow matching for generative modeling.
\newblock \emph{arXiv preprint arXiv:2210.02747}, 2022.

\bibitem[Liu et~al.(2024)Liu, Zhao, Zhuo, Lin, Qiao, Li, and Gao]{luminamgpt}
Dongyang Liu, Shitian Zhao, Le Zhuo, Weifeng Lin, Yu Qiao, Hongsheng Li, and Peng Gao.
\newblock Lumina-mgpt: Illuminate flexible photorealistic text-to-image generation with multimodal generative pretraining.
\newblock \emph{arXiv preprint arXiv:2408.02657}, 2024.

\bibitem[Lu et~al.(2022)Lu, Clark, Zellers, Mottaghi, and Kembhavi]{unifiedio}
Jiasen Lu, Christopher Clark, Rowan Zellers, Roozbeh Mottaghi, and Aniruddha Kembhavi.
\newblock Unified-io: A unified model for vision, language, and multi-modal tasks.
\newblock In \emph{The Eleventh International Conference on Learning Representations}, 2022.

\bibitem[Ma et~al.(2024)Ma, Zhou, Yeh, Wang, Li, Yang, Dong, Keutzer, and Feng]{magicme}
Ze Ma, Daquan Zhou, Chun-Hsiao Yeh, Xue-She Wang, Xiuyu Li, Huanrui Yang, Zhen Dong, Kurt Keutzer, and Jiashi Feng.
\newblock Magic-me: Identity-specific video customized diffusion.
\newblock \emph{arXiv preprint arXiv:2402.09368}, 2024.

\bibitem[Menapace et~al.(2024)Menapace, Siarohin, Skorokhodov, Deyneka, Chen, Kag, Fang, Stoliar, Ricci, Ren, et~al.]{snapvideo}
Willi Menapace, Aliaksandr Siarohin, Ivan Skorokhodov, Ekaterina Deyneka, Tsai-Shien Chen, Anil Kag, Yuwei Fang, Aleksei Stoliar, Elisa Ricci, Jian Ren, et~al.
\newblock Snap video: Scaled spatiotemporal transformers for text-to-video synthesis.
\newblock \emph{arXiv preprint arXiv:2402.14797}, 2024.

\bibitem[Mou et~al.(2024)Mou, Wang, Xie, Wu, Zhang, Qi, and Shan]{t2i_adapter}
Chong Mou, Xintao Wang, Liangbin Xie, Yanze Wu, Jian Zhang, Zhongang Qi, and Ying Shan.
\newblock T2i-adapter: Learning adapters to dig out more controllable ability for text-to-image diffusion models.
\newblock In \emph{Proceedings of the AAAI Conference on Artificial Intelligence}, pages 4296--4304, 2024.

\bibitem[Pan et~al.(2024)Pan, Barath, Pollefeys, and Sch\"{o}nberger]{glomap}
Linfei Pan, Daniel Barath, Marc Pollefeys, and Johannes~Lutz Sch\"{o}nberger.
\newblock {Global Structure-from-Motion Revisited}.
\newblock In \emph{European Conference on Computer Vision (ECCV)}, 2024.

\bibitem[Peebles and Xie(2023)]{dit}
William Peebles and Saining Xie.
\newblock Scalable diffusion models with transformers.
\newblock In \emph{Proceedings of the IEEE/CVF International Conference on Computer Vision}, pages 4195--4205, 2023.

\bibitem[Polyak et~al.(2024)Polyak, Zohar, Brown, Tjandra, Sinha, Lee, Vyas, Shi, Ma, Chuang, et~al.]{moviegen}
Adam Polyak, Amit Zohar, Andrew Brown, Andros Tjandra, Animesh Sinha, Ann Lee, Apoorv Vyas, Bowen Shi, Chih-Yao Ma, Ching-Yao Chuang, et~al.
\newblock Movie gen: A cast of media foundation models.
\newblock \emph{arXiv preprint arXiv:2410.13720}, 2024.

\bibitem[Radford et~al.(2021)Radford, Kim, Hallacy, Ramesh, Goh, Agarwal, Sastry, Askell, Mishkin, Clark, et~al.]{clip}
Alec Radford, Jong~Wook Kim, Chris Hallacy, Aditya Ramesh, Gabriel Goh, Sandhini Agarwal, Girish Sastry, Amanda Askell, Pamela Mishkin, Jack Clark, et~al.
\newblock Learning transferable visual models from natural language supervision.
\newblock In \emph{International conference on machine learning}, pages 8748--8763. PmLR, 2021.

\bibitem[Ruiz et~al.(2023)Ruiz, Li, Jampani, Pritch, Rubinstein, and Aberman]{dreambooth}
Nataniel Ruiz, Yuanzhen Li, Varun Jampani, Yael Pritch, Michael Rubinstein, and Kfir Aberman.
\newblock Dreambooth: Fine tuning text-to-image diffusion models for subject-driven generation.
\newblock In \emph{Proceedings of the IEEE/CVF conference on computer vision and pattern recognition}, pages 22500--22510, 2023.

\bibitem[Schuhmann et~al.(2022)Schuhmann, Beaumont, Vencu, Gordon, Wightman, Cherti, Coombes, Katta, Mullis, Wortsman, et~al.]{laion}
Christoph Schuhmann, Romain Beaumont, Richard Vencu, Cade Gordon, Ross Wightman, Mehdi Cherti, Theo Coombes, Aarush Katta, Clayton Mullis, Mitchell Wortsman, et~al.
\newblock Laion-5b: An open large-scale dataset for training next generation image-text models.
\newblock \emph{Advances in neural information processing systems}, 35:\penalty0 25278--25294, 2022.

\bibitem[Singer et~al.(2022)Singer, Polyak, Hayes, Yin, An, Zhang, Hu, Yang, Ashual, Gafni, et~al.]{makeavideo}
Uriel Singer, Adam Polyak, Thomas Hayes, Xi Yin, Jie An, Songyang Zhang, Qiyuan Hu, Harry Yang, Oron Ashual, Oran Gafni, et~al.
\newblock Make-a-video: Text-to-video generation without text-video data.
\newblock \emph{arXiv preprint arXiv:2209.14792}, 2022.

\bibitem[Skorokhodov et~al.(2022)Skorokhodov, Tulyakov, and Elhoseiny]{mcvd}
Ivan Skorokhodov, Sergey Tulyakov, and Mohamed Elhoseiny.
\newblock Stylegan-v: A continuous video generator with the price, image quality and perks of stylegan2.
\newblock In \emph{Proceedings of the IEEE/CVF Conference on Computer Vision and Pattern Recognition}, pages 3626--3636, 2022.

\bibitem[Team(2024)]{chameleon}
Chameleon Team.
\newblock Chameleon: Mixed-modal early-fusion foundation models.
\newblock \emph{arXiv preprint arXiv:2405.09818}, 2024.

\bibitem[Teed and Deng(2020)]{raft}
Zachary Teed and Jia Deng.
\newblock Raft: Recurrent all-pairs field transforms for optical flow.
\newblock In \emph{Computer Vision--ECCV 2020: 16th European Conference, Glasgow, UK, August 23--28, 2020, Proceedings, Part II 16}, pages 402--419. Springer, 2020.

\bibitem[Vondrick et~al.(2016)Vondrick, Pirsiavash, and Torralba]{vgan}
Carl Vondrick, Hamed Pirsiavash, and Antonio Torralba.
\newblock Generating videos with scene dynamics.
\newblock \emph{Advances in neural information processing systems}, 29, 2016.

\bibitem[Wang et~al.(2024{\natexlab{a}})Wang, Yuan, Zhang, Chen, Wang, Zhang, Shen, Zhao, and Zhou]{videocomposer}
Xiang Wang, Hangjie Yuan, Shiwei Zhang, Dayou Chen, Jiuniu Wang, Yingya Zhang, Yujun Shen, Deli Zhao, and Jingren Zhou.
\newblock Videocomposer: Compositional video synthesis with motion controllability.
\newblock \emph{Advances in Neural Information Processing Systems}, 36, 2024{\natexlab{a}}.

\bibitem[Wang et~al.(2024{\natexlab{b}})Wang, Zhang, Luo, Sun, Cui, Wang, Zhang, Wang, Li, Yu, et~al.]{emu3}
Xinlong Wang, Xiaosong Zhang, Zhengxiong Luo, Quan Sun, Yufeng Cui, Jinsheng Wang, Fan Zhang, Yueze Wang, Zhen Li, Qiying Yu, et~al.
\newblock Emu3: Next-token prediction is all you need.
\newblock \emph{arXiv preprint arXiv:2409.18869}, 2024{\natexlab{b}}.

\bibitem[Wang et~al.(2024{\natexlab{c}})Wang, Yuan, Wang, Li, Chen, Xia, Luo, and Shan]{motionctrl}
Zhouxia Wang, Ziyang Yuan, Xintao Wang, Yaowei Li, Tianshui Chen, Menghan Xia, Ping Luo, and Ying Shan.
\newblock Motionctrl: A unified and flexible motion controller for video generation.
\newblock In \emph{ACM SIGGRAPH 2024 Conference Papers}, pages 1--11, 2024{\natexlab{c}}.

\bibitem[Wei et~al.(2024)Wei, Zhang, Qing, Yuan, Liu, Liu, Zhang, Zhou, and Shan]{dreamvideo}
Yujie Wei, Shiwei Zhang, Zhiwu Qing, Hangjie Yuan, Zhiheng Liu, Yu Liu, Yingya Zhang, Jingren Zhou, and Hongming Shan.
\newblock Dreamvideo: Composing your dream videos with customized subject and motion.
\newblock In \emph{Proceedings of the IEEE/CVF Conference on Computer Vision and Pattern Recognition}, pages 6537--6549, 2024.

\bibitem[Wu et~al.(2024{\natexlab{a}})Wu, Chen, Wu, Ma, Liu, Pan, Liu, Xie, Yu, Ruan, et~al.]{janus}
Chengyue Wu, Xiaokang Chen, Zhiyu Wu, Yiyang Ma, Xingchao Liu, Zizheng Pan, Wen Liu, Zhenda Xie, Xingkai Yu, Chong Ruan, et~al.
\newblock Janus: Decoupling visual encoding for unified multimodal understanding and generation.
\newblock \emph{arXiv preprint arXiv:2410.13848}, 2024{\natexlab{a}}.

\bibitem[Wu et~al.(2024{\natexlab{b}})Wu, Li, Zeng, Zhang, Zhou, Li, Tong, and Chen]{motionbooth}
Jianzong Wu, Xiangtai Li, Yanhong Zeng, Jiangning Zhang, Qianyu Zhou, Yining Li, Yunhai Tong, and Kai Chen.
\newblock Motionbooth: Motion-aware customized text-to-video generation.
\newblock \emph{arXiv preprint arXiv:2406.17758}, 2024{\natexlab{b}}.

\bibitem[Xiao et~al.(2024)Xiao, Wang, Zhou, Yuan, Xing, Yan, Wang, Huang, and Liu]{omnigen}
Shitao Xiao, Yueze Wang, Junjie Zhou, Huaying Yuan, Xingrun Xing, Ruiran Yan, Shuting Wang, Tiejun Huang, and Zheng Liu.
\newblock Omnigen: Unified image generation.
\newblock \emph{arXiv preprint arXiv:2409.11340}, 2024.

\bibitem[Xie et~al.(2024)Xie, Mao, Bai, Zhang, Wang, Lin, Gu, Chen, Yang, and Shou]{showo}
Jinheng Xie, Weijia Mao, Zechen Bai, David~Junhao Zhang, Weihao Wang, Kevin~Qinghong Lin, Yuchao Gu, Zhijie Chen, Zhenheng Yang, and Mike~Zheng Shou.
\newblock Show-o: One single transformer to unify multimodal understanding and generation.
\newblock \emph{arXiv preprint arXiv:2408.12528}, 2024.

\bibitem[Xie et~al.(2023)Xie, Jampani, Zhong, Sun, and Jiang]{omnicontrol}
Yiming Xie, Varun Jampani, Lei Zhong, Deqing Sun, and Huaizu Jiang.
\newblock Omnicontrol: Control any joint at any time for human motion generation.
\newblock \emph{arXiv preprint arXiv:2310.08580}, 2023.

\bibitem[Xu et~al.(2024)Xu, He, Shan, and Chen]{ctrlora}
Yifeng Xu, Zhenliang He, Shiguang Shan, and Xilin Chen.
\newblock Ctrlora: An extensible and efficient framework for controllable image generation.
\newblock \emph{arXiv preprint arXiv:2410.09400}, 2024.

\bibitem[Yang et~al.(2024)Yang, Kang, Huang, Xu, Feng, and Zhao]{depthanything}
Lihe Yang, Bingyi Kang, Zilong Huang, Xiaogang Xu, Jiashi Feng, and Hengshuang Zhao.
\newblock Depth anything: Unleashing the power of large-scale unlabeled data.
\newblock In \emph{Proceedings of the IEEE/CVF Conference on Computer Vision and Pattern Recognition}, pages 10371--10381, 2024.

\bibitem[Yuan et~al.(2024)Yuan, Huang, He, Ge, Shi, Chen, Luo, and Yuan]{consisid}
Shenghai Yuan, Jinfa Huang, Xianyi He, Yunyuan Ge, Yujun Shi, Liuhan Chen, Jiebo Luo, and Li Yuan.
\newblock Identity-preserving text-to-video generation by frequency decomposition.
\newblock \emph{arXiv preprint arXiv:2411.17440}, 2024.

\bibitem[Zhang et~al.(2024)Zhang, Li, Le, Shou, Xiong, and Sahoo]{moonshot}
David~Junhao Zhang, Dongxu Li, Hung Le, Mike~Zheng Shou, Caiming Xiong, and Doyen Sahoo.
\newblock Moonshot: Towards controllable video generation and editing with multimodal conditions.
\newblock \emph{arXiv preprint arXiv:2401.01827}, 2024.

\bibitem[Zhang et~al.(2023{\natexlab{a}})Zhang, Rao, and Agrawala]{controlnet}
Lvmin Zhang, Anyi Rao, and Maneesh Agrawala.
\newblock Adding conditional control to text-to-image diffusion models.
\newblock In \emph{Proceedings of the IEEE/CVF International Conference on Computer Vision}, pages 3836--3847, 2023{\natexlab{a}}.

\bibitem[Zhang et~al.(2023{\natexlab{b}})Zhang, Wei, Jiang, Zhang, Zuo, and Tian]{controlvideo}
Yabo Zhang, Yuxiang Wei, Dongsheng Jiang, Xiaopeng Zhang, Wangmeng Zuo, and Qi Tian.
\newblock Controlvideo: Training-free controllable text-to-video generation.
\newblock \emph{arXiv preprint arXiv:2305.13077}, 2023{\natexlab{b}}.

\bibitem[Zhang et~al.(2025)Zhang, Liu, Xia, Peng, Yan, Lo, and Jia]{MagicMirror}
Yuechen Zhang, Yaoyang Liu, Bin Xia, Bohao Peng, Zexin Yan, Eric Lo, and Jiaya Jia.
\newblock Magic mirror: Id-preserved video generation in video diffusion transformers.
\newblock \emph{arXiv preprint arXiv:2501.03931}, 2025.

\bibitem[Zheng et~al.(2024)Zheng, Li, Jiang, Lu, Wu, and Li]{cami2v}
Guangcong Zheng, Teng Li, Rui Jiang, Yehao Lu, Tao Wu, and Xi Li.
\newblock Cami2v: Camera-controlled image-to-video diffusion model.
\newblock \emph{arXiv preprint arXiv:2410.15957}, 2024.

\bibitem[Zhou et~al.(2024{\natexlab{a}})Zhou, Yu, Babu, Tirumala, Yasunaga, Shamis, Kahn, Ma, Zettlemoyer, and Levy]{transfusion}
Chunting Zhou, Lili Yu, Arun Babu, Kushal Tirumala, Michihiro Yasunaga, Leonid Shamis, Jacob Kahn, Xuezhe Ma, Luke Zettlemoyer, and Omer Levy.
\newblock Transfusion: Predict the next token and diffuse images with one multi-modal model.
\newblock \emph{arXiv preprint arXiv:2408.11039}, 2024{\natexlab{a}}.

\bibitem[Zhou et~al.(2024{\natexlab{b}})Zhou, Wang, Nie, Lin, Yu, Yu, and Wang]{trackgo}
Haitao Zhou, Chuang Wang, Rui Nie, Jinxiao Lin, Dongdong Yu, Qian Yu, and Changhu Wang.
\newblock Trackgo: A flexible and efficient method for controllable video generation.
\newblock \emph{arXiv preprint arXiv:2408.11475}, 2024{\natexlab{b}}.

\bibitem[Zhou et~al.(2018)Zhou, Tucker, Flynn, Fyffe, and Snavely]{realestate10k}
Tinghui Zhou, Richard Tucker, John Flynn, Graham Fyffe, and Noah Snavely.
\newblock Stereo magnification: Learning view synthesis using multiplane images.
\newblock \emph{arXiv preprint arXiv:1805.09817}, 2018.

\end{thebibliography}
}

\clearpage
\appendix

\end{document}